\documentclass[sigconf]{acmart}

\usepackage{fancyhdr}
\usepackage{algorithmic}
\usepackage{graphicx}
\usepackage{textcomp}
\usepackage{xcolor}
\usepackage{bm}
\usepackage{booktabs}
\usepackage[ruled,linesnumbered]{algorithm2e}
\usepackage{multirow}
\usepackage{enumitem}
\usepackage{microtype}
\usepackage{subfigure}
\usepackage{booktabs} 
\usepackage{amsfonts} 
\usepackage{balance}

\usepackage{amssymb}
\usepackage{amsmath}
\usepackage{sansmath}
\usepackage{subfigure}
\usepackage{amsthm}
\usepackage{tcolorbox}
\usepackage{colortbl}
\usepackage{tikz}
\usepackage{url}
\usepackage{pifont}
\usepackage{makecell}
\usepackage{cleveref}
\usepackage{pgfplots}
\usepackage{setspace}
\usepackage{xcolor}

\crefformat{section}{\S#2#1#3} 
\crefformat{subsection}{\S#2#1#3}
\long\def\comment#1{}

\newcommand{\PtoP}{{\textsc{P2P}}\xspace}
\newcommand{\MBTI}{\text{MBTI}\xspace}
\newcommand{\nthesection}{\arabic{section}}

\newcounter{remark}[section]
\renewcommand{\theremark}{\nthesection.\arabic{remark}}

\newcommand{\stitle}[1]{\vspace{1ex} \noindent{\bf #1}}

\newcommand{\beqn}{\begin{eqnarray*}}
\newcommand{\eeqn}{\end{eqnarray*}}

\newcounter{ccc}

\newcommand{\cmark}{\ding{51}}%
\newcommand{\ccross}{\ding{56}}

\AtBeginDocument{%
  }

\copyrightyear{2025} 
\acmYear{2025} 
\setcopyright{acmlicensed}\acmConference[CIKM '25]{Proceedings of the 34th ACM International Conference on Information and Knowledge Management}{November 10--14, 2025}{Seoul, Republic of Korea}
\acmBooktitle{Proceedings of the 34th ACM International Conference on Information and Knowledge Management (CIKM '25), November 10--14, 2025, Seoul, Republic of Korea}
\acmDOI{10.1145/3746252.3760813}
\acmISBN{979-8-4007-2040-6/2025/11}

\begin{document}

\title{From Post To Personality: Harnessing LLMs for MBTI Prediction in Social Media}


\author{Tian Ma}
\affiliation{%
 \institution{Beijing Institute of Technology}
  \city{Beijing}
 \country{China}
}

\author{Kaiyu Feng}
\affiliation{%
  \institution{Beijing Institute of Technology}
  \city{Beijing}
  \country{China}
  }

\author{Yu Rong}
\affiliation{%
  \institution{The Chinese University of Hong Kong}
  \city{Hong Kong S. A. R.}
  \country{China}
  }

\author{Kangfei Zhao}
\authornote{Corresponding author. Email: zkf1105@gmail.com}
\affiliation{%
  \institution{Beijing Institute of Technology}
  \city{Beijing}
  \country{China}
  }

\renewcommand{\shortauthors}{Tian Ma, Kaiyu Feng, Yu Rong and Kangfei Zhao}

\begin{abstract}
Personality prediction from social media posts is a critical task that implies diverse applications in psychology and sociology.
The Myers–Briggs Type Indicator (\MBTI), a popular personality inventory, has been traditionally predicted by machine learning (ML) and deep learning (DL) techniques.
Recently, the success of Large Language Models (LLMs) has revealed their huge potential in understanding and inferring personality traits from social media content. 
However, directly exploiting LLMs for \MBTI prediction faces two key challenges: the hallucination problem inherent in  LLMs and the naturally imbalanced distribution of \MBTI types in the population. 
In this paper, we propose PostToPersonality (\PtoP), a novel LLM-based framework for \MBTI prediction from social media posts of individuals. 
Specifically, \PtoP leverages Retrieval-Augmented Generation with in-context learning to mitigate hallucination in LLMs. Furthermore, we fine-tune a pretrained LLM to improve model specification in \MBTI understanding with synthetic minority oversampling, which balances the class imbalance by generating synthetic samples. Experiments conducted on a real-world social media dataset demonstrate that \PtoP achieves state-of-the-art performance compared with 10 ML/DL baselines.
\end{abstract}




\maketitle

\section{Introduction}
Predicting the personality characteristics of social media users is an intriguing task for social media text mining, underpinning various applications such as social recommendation~\cite{feng2013recommendation, dhelim2020personality}, educational and career planning~\cite{hogan1996personality}, and online psychosocial assessment~\cite{saha2020psychosocial}. 
Myers-Briggs Type Indicator (\MBTI)~\cite{myers1962myers} is a popular personality trait, especially for young people active in social media.
It is estimated that in the U.S., millions of people have taken the \MBTI and thousands of commercial and educational institutions use \MBTI~\cite{seattletimes}. 
The \MBTI categorizes individuals into 16 personality types along four dimensions: Extraversion vs.
Introversion, Sensing vs. Intuition, Thinking vs. Feeling and Judging vs. Perceiving. 


In past years, machine learning (ML) techniques have been used 
to predict the \MBTI from an individual's textual input. 
Various classifiers~\cite{Garg2021,Choong2021,Khan2020,Abidin2020,Das2020,Mushtaq2020} are constructed with TF-IDF or bag-of-words as the text features. 
These approaches heavily rely on feature engineering and yield suboptimal accuracy due to the lack of end-to-end optimization.
Deep learning (DL) approaches such as BERT and Transformer alleviate these limitations by automatic feature extraction and contextual understanding. However, these models only possess a limited perception of nuanced social and psychological characteristics for \MBTI prediction.

Recently, we have witnessed the great potential of Large Language Models (LLMs)~\cite{DBLP:conf/www/LuceriBF24, DBLP:journals/snam/ThapaSSAVNN25} in social media analytics. Pretrained on vast corpora, LLMs exhibit surprising emergent abilities for understanding human sentiment and personality traits. Furthermore, when fine-tuned with social media data, these models can leverage in-context information to infer the behaviors and preferences of users. 
Despite the promise, directly deploying or fine-tuning LLMs for \MBTI prediction from social posts faces two intertwined challenges. 
First, general-purpose LLMs struggle with the hallucination problem, probably leading to inaccurate personality assessments.
Second, the training data of social posts with \MBTI ground-truth is scarce, and the distribution of the 16 types of personality, following the underlying distribution in the population, is highly imbalanced.

To this end, we propose an innovative LLM-based framework, named PostToPersonality (\PtoP), which predicts the \MBTI personality from social media posts. 
In a nutshell, \PtoP first generates a high-level psychological assessment of user posts by a locally deployed LLM, and then combines this assessment with the original posts to predict \MBTI types via an online LLM. 
This dual-LLM architecture incorporates flexible adaptation of LLM to \MBTI prediction and the advanced reasoning capability of online LLMs. 
The contributions of this paper are summarized as follows: 
\begin{itemize}[leftmargin=*]
\item We propose an LLM-based framework that predicts the \MBTI personality from individuals' posts in social media. The framework, named \PtoP, integrates the powerful reasoning capability of online LLM and the customization of a local LLM. 
\item \PtoP employs Retrieval-Augmented Generation (RAG) mechanism to alleviate the hallucination of the LLM. Moreover, to fulfill effective customization, we fine-tune the local LLM with synthetic minority over-sampling to resist the imbalance of training data. 
\item We conduct experimental studies on a real-world dataset, which demonstrates that \PtoP outperforms 10 ML/DL baselines by an average of 8.2\% in accuracy, 20.17\% in F1-score, and 4.1\% in AUC. Further case studies validate the potential of LLMs in capturing personality characteristics from social posts. 
\end{itemize}


\stitle{Related Works.}
Early research in \MBTI prediction began with Argamon et al. \cite{Argamon2005}  who used SVMs to analyze student essays by categorizing words into functional types.  
Amirhosseini and Kazemian~\cite{Amirhosseini2020} introduced TF-IDF and XGBoost for MBTI, while Gjurković et al.~ \cite{Gjurkovic2021} leveraged neural networks, achieving F1 scores between 63.4\% and 73.9\%. 
Subsequent studies explored various classifiers~\cite{Garg2021,Choong2021,Khan2020,Abidin2020,Das2020,Mushtaq2020}, including MLP and Random Forest, for \MBTI classification tasks. 
DL efforts, Cui and Qi \cite{Cui2017} reported only 23\% accuracy with LSTM. 


Recent work has expanded personality analysis to LLMs, with some studies maintaining an MBTI focus while others examining alternative frameworks. Pan et al. \cite{Pan2023} specifically investigated whether LLMs exhibit MBTI personality traits by administering the MBTI test and attempting to modify their MBTI attributes. Similarly, Rao et al. \cite{Rao2023} evaluated open-source LLMs' ability to mimic human MBTI profiles. In contrast, other researchers have explored different personality models, with Yan et al. \cite{yan2024predictingbigpersonalitytraits} focusing on Big Five personality traits prediction from psycho-counseling dialogues, and Wang et al. \cite{wang2025evaluating} assessing GPT-4's role-play ability based on Big Five profiles rather than MBTI characteristics.

 \section{Preliminary \& Problem Statement}
 \label{sec:preliminary}

The Myers–Briggs Type Indicator (\MBTI) categorizes individuals into 16 personality types along  four dimensions: 
\ding{172} Extraversion (E) vs. Introversion (I) 
\ding{173} Sensing (S) vs. Intuition (N)
\ding{174} Thinking (T) vs. Feeling (F)
\ding{175} Judging (J) vs. Perceiving (P).
Each personality type is identified by a unique combination of four characters $\bm{c} = c_1 \circ c_2 \circ c_3 \circ c_4$ in these dimensions, where $c_1 \in \{E, I\}$, $c_2 \in \{S, N\}$, $c_3 \in \{T, F\}$, $c_4 \in \{J, P\}$, and $\circ$ denotes concatenation. 
Given a list of social media posts of a user, we aim to build a framework based on open-source and online LLMs to directly predict the personality type of the user, where the prediction task is formulated as four binary classification tasks. Let $\bm{x}$ denote the historical posts of a user, $\bm{p}$ denote the prompt,  $\bm{h}$ denote possible textual features extracted from these posts, and $\bm{d}$ denote potential demonstrations, respectively. We build a learning framework $\pi_{\theta}: (\bm{p}, \bm{x}, \bm{h}, \bm{d}) \mapsto \bm{c}$ that predicts the \MBTI personality $\bm{c}$ of the user.

\section{P2P: Methodology}
\label{sec:method}

\begin{figure}[t!]
    \centering
    \includegraphics[width=0.48\textwidth]{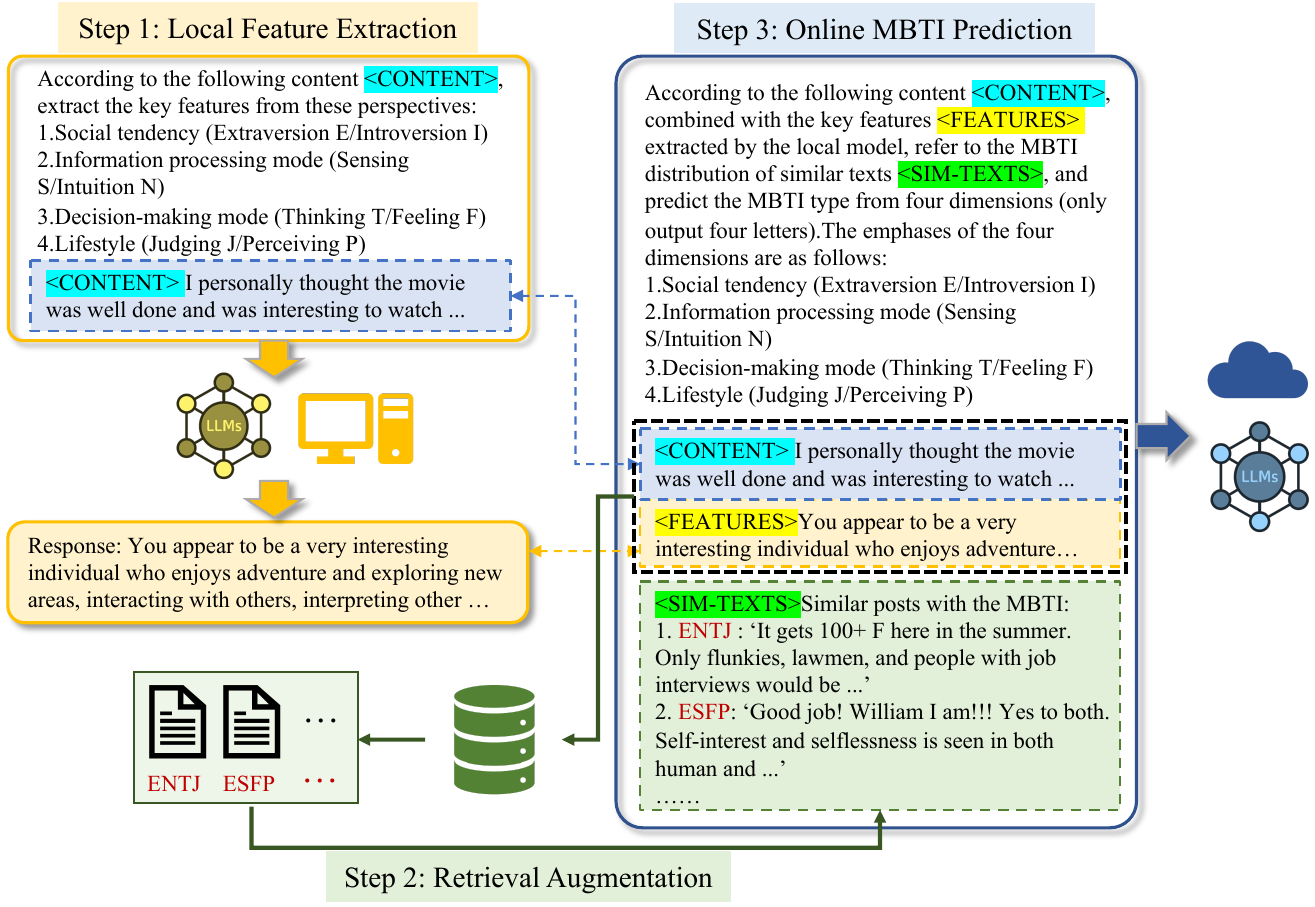} 
    \vspace{-2ex}
    \caption{The Overview of PostToPersonality}
    \label{fig:prompt}
\vspace{-0.1cm}
\end{figure}

Fig.~\ref{fig:prompt} illustrates an overview of the \PtoP framework, which predicts the personality of a user from her posts in three sequential steps, leveraging a fine-tuned local LLM, a post-personality vector database, and online LLM calling. 
In step 1, \PtoP generates interpretable psychological assessments that capture key personality features from the raw posts using a fine-tuned local LLM. 
Subsequently, in step 2, the generated textual features are paired with the original posts, and \PtoP uses their embeddings as the key to query the post-personality vector database. 
This retrieval process identifies the top-$k$ most semantically similar historical posts, along with their ground-truth \MBTI labels, thereby providing rich contextual personality demonstrations.
Finally, in step 3, \PtoP synthesizes a comprehensive prompt and issues an online LLM call to predict the \MBTI. The prompt comprises the original posts, the generated textual features from local LLMs, and the retrieved $k$ post-personality demonstrations. This multi-step dual-LLM framework exploits a steerable fine-tuned LLM for task-specific feature augmentation, and simultaneously harnesses the reasoning and generalization power of commercial online LLMs. 
In the following, we will elaborate on the online prediction in \cref{sec:method:prediction} and introduce the fine-tuning of the local LLM in \cref{sec:method:finetune}.

\subsection{Online Prediction with RAG}
\label{sec:method:prediction}

\PtoP employs Retrieval-Augmented Generation (RAG) to enhance \MBTI prediction by an online commercial LLM, which retrieves semantically similar post-personality pairs as demonstrations $\bm{d}$. 
As a preparatory offline step, we construct a vector database $\mathcal{D}$ to index collected posts alongside their textual features generated by the fine-tuned local LLM. 
Specifically, for a user, the original posts, concatenated in a long text, are encoded into vector embeddings using Sentence-Bert~\cite{DBLP:conf/emnlp/ReimersG19}, while the textual features are encoded into embeddings by the local LLM. The two embeddings are transformed into two vectors by average pooling, respectively, which are subsequently concatenated into a single vector. 
\PtoP employs FAISS~\cite{DBLP:journals/tbd/JohnsonDJ21} to build the vector database $\mathcal{D}$, which is persisted for efficient retrieval. 


During the prediction stage, given a user's posts $\bm{x}$, as the textual features $\bm{h}$ are generated from the local LLM, we generate the query vector following a similar way to the construction of the database $\mathcal{D}$. 
Subsequently, \PtoP uses the query vector to retrieve the top-$k$ most similar entries in $\mathcal{D}$, where their corresponding posts $\bm{x}_i$ and the ground-truth \MBTI label $c_i$ for $i \in [1, k]$ are collected as the demonstration set $\bm{d} = \{(\bm{x}'_i, \bm{c}'_i) \mid i \in [1, k] \}$. 
As illustrated in Fig.~\ref{fig:prompt}, \PtoP combines the posts $\bm{x}$, the generated textual features $\bm{h}$ and the demonstrations $\bm{h}$ into one prompt and submits the prompt to an online LLM to predict the \MBTI for $\bm{x}$ by in-context learning~\cite{dong-etal-2024-survey}.

\subsection{Finetuning Local LLM}
\label{sec:method:finetune}

\PtoP deploys an open-source LLM locally to summarize and synthesize salient textual features of a user's posts as an interpretable assessment. 
To enable a general-purpose LLM to effectively understand personality-related information, we fine-tune the LLM by a supervised learning task that directly predicts the \MBTI type from the posts. Let $p_{\theta}(\bm{c} \mid \bm{x})$ denote the predictive conditional probability of the \MBTI $\bm{c}$ given the posts $\bm{x}$, we minimize the negative log-likelihood of this probability over a training set, as formalized in Eq.~\eqref{eq:loss:sft}, by stochastic gradient descent. 
\begin{align}
\label{eq:loss:sft}
\mathcal{L}_{\text{SFT}}(\theta) &= \mathop{\mathbb{E}}_{(\bm{x},\bm{c})\sim\mathbb{D}} -\log p_{\theta}(\bm{c}\mid\bm{x})
\end{align}
This fine-tuning is implemented by QA-LoRA~\cite{DBLP:conf/iclr/XuXG0CZC0024}, a performance-efficient fine-tuning algorithm for LLMs.

A critical challenge in the fine-tuning is the out-of-distribution (OOD) problem, stemming from the highly imbalanced distribution of \MBTI types in the training data due to its underlying distribution in the population.
This imbalance can lead to poor predictive performance on minority MBTI types, such as ESFJ and ESTJ. 
Therefore, we enhance the fine-tuning process with the Synthetic Minority Over-sampling Technique (SMOTE)~\cite{smote}, which mitigates the OOD issue by generating synthetic samples for underrepresented classes.
Let $\mathbf{x}_i$ denote the hidden representation in the final layer of the local LLM for post $\bm{x}_i$, and $\overline{\mathbf{x}}_i$ denote the average pooling result of $\mathbf{x}_i$. For $\mathbf{x}_i$ in minority \MBTI types, SMOTE synthesizes an embedding $\hat{\mathbf{x}}_i$ by linearly interpolating between $\mathbf{x}_i$ and a sample $\mathbf{x}_j$ drawn from the neighbor set of $\mathbf{x}_i$, as shown in Eq.~\eqref{eq:smote}:
\begin{align}
    \hat{\mathbf{x}_i} &= \mathbf{x}_i + \lambda ( \mathbf{x}_i - \mathbf{x}_j), \lambda \sim \text{Uniform}(0, 1), \mathbf{x}_j \sim \mathcal{N}(\mathbf{x}_i) \label{eq:smote}\\
    \mathcal{N}(\mathbf{x}_i) &= \arg \min d(\overline{\mathbf{x}}_i, \overline{\mathbf{x}}_j), \ s.t. \ \bm{c}_i = \bm{c}_j
\end{align}
A neighbor of $\mathbf{x}_i$, $\mathbf{x}_j$, is the hidden representation of posts sharing the same \MBTI label with $\bm{x}_i$, and the distance between $\overline{\mathbf{x}}_i$ and $\overline{\mathbf{x}}_i$ is the top-$k$ smallest. Here, the distance $d(\overline{\mathbf{x}}_i, \overline{\mathbf{x}}_j)$ function is a weighted sum of the L2 distance and the cosine distance, and \MBTI types except the most-frequent type (i.e., {INFP}) are treated as minority classes.  
The synthetic sample $\hat{\mathbf{x}}_i$ is added to the training set for local LLM fine-tuning, thereby improving the robustness and generalization of the model. 

\begin{table*}[t]
\caption{Overall Performance (\text{The first and second best scores are in {\bf bold} and \underline{underlined}.)}}
\label{tab:contrast}
\vspace{-3ex}
\centering
\footnotesize
\begin{tabular}{c|ccc|ccc|ccc|ccc}
\toprule
\multirow{2}{*}{Approach} & \multicolumn{3}{c|}{I/E} & \multicolumn{3}{c|}{N/S} & \multicolumn{3}{c|}{T/F} & \multicolumn{3}{c}{J/P} \\ 
& Acc & F1 & AUC & Acc & F1 & AUC & Acc & F1 & AUC & Acc & F1 & AUC  \\ \midrule
Naive Bayes &0.6853 & 0.3374&0.5932 & 0.7337&0.2759 &0.5833 &0.7072 &0.7008 &0.7204 & 0.5954& 0.4674&0.5775 \\ 
LR & 0.8583&0.6192 &0.8938 &0.8963 &0.4601 & 0.9036&0.8503 &0.8325 &\underline{0.9292} & 0.8087&0.7432 &0.8868 \\ 
SVM & 0.8609& 0.6261& 0.8940&0.9029 & 0.5142& \textbf{0.9184} &\underline{0.8541} & \underline{0.8369}&0.9251 &0.8077 &0.7379 &0.8867  \\
XGBoost &\underline{0.8672} & 0.6810&\underline{0.9022} &\underline{0.9087} & 0.6025&\underline{0.9098} &0.8441 &0.8285 &0.9204 & \underline{0.8182}& \underline{0.7619}&\underline{0.8966}  \\
MLP &0.8201 &0.5738 & 0.8344&0.8905 &0.4993 &0.8572 &0.8035 &0.7858 &0.8901 &0.7442 & 0.6739&0.8162  \\
Word2Vec &0.7196 &\underline{0.8172} &0.6801 & 0.6785&0.4349 &0.6832 &0.6785 & 0.6379&0.7362 &0.6525 &0.7044 &0.7047 \\
e5-base &0.7850 &0.1853 &0.5193 &0.4882 &0.2845 & 0.4393& 0.5320&0.4226 & 0.4716&0.5648 & 0.6084&0.6123  \\ 
ModernBERT-base & 0.8519&0.6226 &0.8469 &0.9014& \underline{0.6307}&0.8947 &0.7954 & 0.7963& 0.8890& 0.4916&0.4821 &0.4535 \\ 
DeBERTa-v3-base & 0.7689&0.4252 &0.5110 &0.8617& 0.4721&0.5278 &0.5412 & 0.4978& 0.5175& 0.6044&0.5673 &0.5927 \\ 
DeepSeek-8B &  0.7839 & 0.6939& 0.6395&0.4181 &0.3052 &0.4912 & 0.5401&0.3788 & 0.5010& 0.5836&0.4126 &0.4994  \\ 
\midrule
P2P (ours) &\textbf{0.9321}&\textbf{0.9224} & \textbf{0.9574} & \textbf{0.9475} & \textbf{0.8747} & 0.8849 & \textbf{0.9306} & \textbf{0.9342} & \textbf{0.9602} & \textbf{0.8858} & \textbf{0.8759} & \textbf{0.9059}  \\
\bottomrule   
\end{tabular}
\end{table*}

\section{Experimental Studies}
\label{sec:exp}
We give the test settings in~\cref{sec:exp:setup} and report our substantial experiments in the following facets:
\ding{172} Compare the effectiveness of \PtoP with baselines (\cref{sec:exp:effectiveness})
\ding{173} Ablate the design options of \PtoP (\cref{sec:exp:sensitivity})
\ding{174} Conduct case studies for \MBTI prediction (\cref{sec:exp:casestudy}). 

\subsection{Experimental Setup}
\label{sec:exp:setup}
\stitle{Dataset.}
We use the dataset from \textit{PersonalityCafe} forum~\cite{pcdata} to evaluate our framework, whose statistical profiles are illustrated in Fig.~\ref{fig:distribution}. The dataset comprises posts and \MBTI labels of 8,675 users. Each user is associated with her/his recent 50 posts, which are concatenated into a long text by a special token. 
%
To ensure data consistency, we preprocess the dataset by converting all text to lowercase, performing word lemmatization, and removing stopwords, special characters, links, URLs, and extra whitespace.
We split the dataset by 60\% for training, 20\% for validation, and 20\% for testing.


\stitle{Baselines.} 
We compare P2P against 10 baselines:
\ding{172} Multi-nomial Naive Bayes Classifier
\ding{173} Logistic Regression (LR)
\ding{174} Support Vector Classifier (SVM)~\cite{vapnik1963generalized}
\ding{175} XGBoost~\cite{DBLP:conf/kdd/ChenG16}
\ding{176} Multi-layer Perceptron Classifier (MLP)
\ding{177} Word2Vec~\cite{DBLP:journals/corr/abs-1301-3781}
\ding{178} e5-base~\cite{DBLP:journals/corr/abs-2212-03533}
\ding{179} ModernBERT-base~\cite{modernbert}
\ding{180} DeBERTa-v3-base~\cite{he2021debertav3}
\ding{181} DeepSeek-R1-8B~\cite{deepseekai2025deepseekr1incentivizingreasoningcapability, DS8B}.

\stitle{Implementation Details.}
The learning framework is built on PyTorch. We use DeepSeek-R1-8B (DS) as the local LLM and utilize DeepSeek API (DS-V3) for online prediction. 
The local LLM is fine-tuned by QA-LoRA~\cite{DBLP:conf/iclr/XuXG0CZC0024} with the AdamW~\cite{DBLP:conf/iclr/LoshchilovH19} optimizer in 10 epochs, with a learning rate of 1e-4 and a batch size of 8. We use early stopping and a dropout ratio of 5\% to prevent overfitting. 
The vector database is constructed from the training data of the local LLM. 
\PtoP uses FAISS~\cite{DBLP:journals/tbd/JohnsonDJ21} to perform exact $k$-NN search, with $k =5$ and L2 distance as the similarity metric by default. 
The implementation of baselines can be found in our Appendix~\ref{sec:appendix:baselines}.

\stitle{Evaluation Metrics.}
We report Accuracy, F1-score, and Area Under the Curve (AUC) as evaluation metrics.
For \PtoP, the F1 and Accuracy are computed from the final results of the API call, while predictive probabilities for computing AUC are approximated by the retrieved $k$ samples in the RAG. Predictive probability is approximated by the distribution of the ground-truth labels of these samples with appropriate smoothing. 

\subsection{Experimental Results}
\subsubsection{Effectiveness}
\label{sec:exp:effectiveness}

We compare the test Accuracy (Acc), F1 score, and AUC of \PtoP with the 10 baselines. Table \ref{tab:contrast} summarizes the results across the four \MBTI personality dimensions. 
On average, \PtoP achieves an F1-score improvement of 20.17\% over the second-best baseline.
As a leading ML approach, XGBoost achieves strong performance primarily through precise identification of highly discriminative vocabularies. 
For example, Thinking-type (T) users often use abstract terms such as `personally' and `learning', while Feeling-type (F) users tend to favor emotional expressions like  `funny' and `relaxing'. 
DL models such as ModernBERT struggle with the insufficient and long-tail distribution of training data. 
All models exhibit relatively lower performance in distinguishing between Judging (J) and Perceiving (P) types, primarily due to the inherent ambiguity in behavioral traits. The `planning-oriented' nature of J-type and the `spontaneous adaptability' of P-type often overlap in social media posts.

\subsubsection{Ablation Studies}
\label{sec:exp:sensitivity}

\begin{table*}[t!]
\footnotesize 
\caption{Ablation studies for PEFT, SMOTE, RAG, different local/online LLMs and different $k$}
\vspace{-0.1cm}
\label{tab:ablation}
\vspace{-2ex}
\centering
\begin{tabular}{cccccc|ccc|ccc|ccc|ccc}
\toprule
\multirow{2}{*}{Variant} & \multirow{2}{*}{LLMs} & \multirow{2}{*}{PEFT} & \multirow{2}{*}{SMOTE} & \multirow{2}{*}{RAG} & \multirow{2}{*}{k} & \multicolumn{3}{c|}{I/E} & \multicolumn{3}{c|}{N/S}  & \multicolumn{3}{c|}{T/F} & \multicolumn{3}{c}{J/P} \\ 
& & & & & & Acc &  F1 & AUC & Acc & F1 & AUC & Acc & F1 & AUC & Acc & F1 & AUC \\ \midrule
(a) & DS/DS & \cmark & \cmark & \cmark & 5  &0.9321&0.9224 &0.9574 & 0.9475& 0.8747&0.8849 &0.9306 &0.9342 &0.9602 &0.8858 &0.8759 & 0.9059\\ \midrule
(b) & DS/DS & \cmark & \ccross & \cmark & 5  &0.9097 &0.9051 &0.9651 &0.8894 &0.8702 &0.8906 & 0.9414& 0.9530&0.9516 &0.8646 & 0.8723& 0.8443\\
(c) & DS/DS & \cmark & \cmark & \ccross & - &0.8287 &0.7304 & 0.7867&0.8619 &0.6575 & 0.7727&0.8564 & 0.8632& 0.8199& 0.8287&0.7669 & 0.7934\\
(d) & DS/DS & \cmark & \ccross & \ccross & -  &0.7839 & 0.6939& 0.6395&0.4181 &0.3052 &0.4912 & 0.5401&0.3788 & 0.5010& 0.5836&0.4126 &0.4994  \\
(e) & DS/DS & \ccross & - & \cmark & 5  & 0.8950& 0.8155&0.8611 & 0.9061&0.7792 & 0.8242& 0.9006& 0.9053&0.8826 &0.8674 &0.8261 & 0.8547\\
(f) & DS/DS & \ccross & - & \ccross & - & 0.7917& 0.8000& 0.8528& 0.8750&0.8750 & 0.8997&0.8125 & 0.8138& 0.7991&0.7986 &0.7642 &0.8075 \\ \midrule
(g) & MC/DS& \cmark & \cmark& \cmark & 5  & 0.9061& 0.8321&0.8493 &0.9329 &0.7667 &0.8156 & 0.9265&0.9159 &0.9173 & 0.9102& 0.8824&0.8297  \\
(h) & GLM/DS & \cmark & \cmark & \cmark & 5  &0.8964 &0.8951 &0.9409 &0.8672 & 0.8527&0.8783 & 0.9383& 0.9359& 0.9243& 0.8403&0.8296 &0.8254 \\ 
(i) & DS/GPT-4  &  \cmark & \cmark & \cmark & 5 &0.9088 & 0.8137& 0.8464&0.9340 & 0.7791& 0.8510& 0.9124&0.9001 & 0.8824 & 0.8812&0.8465 &0.8001 \\
\midrule
(j) & DS/DS  & \cmark & \cmark & \cmark & 2 &0.9031 &0.7941 &0.8833 & 0.9383&0.7617 & 0.8536& 0.8858&0.8750 &0.8969 & 0.8627&0.8102 & 0.8107\\
(k) & DS/DS  & \cmark & \cmark & \cmark & 3 &0.9026 &0.8986 &0.9302 & 0.8611& 0.8437&0.8912 &0.9258 &0.9413 &0.9574 & 0.8194&0.7969 &0.8663 \\
(l) & DS/DS  &  \cmark & \cmark & \cmark & 4 &0.8658 & 0.8025& 0.8703& 0.8472& 0.7901&0.8511 &0.8947 & 0.8734&0.8852 & 0.8656&0.8132 &0.7901 \\
(m) & DS/DS  &  \cmark & \cmark & \cmark & 6 &0.9306 & 0.9213& 0.9728&0.8681 & 0.8504& 0.8625& 0.9428&0.9517 & 0.9475& 0.8403&0.8244 &0.8061 \\
\bottomrule  
\end{tabular}
\vspace{-2ex}
\end{table*}
We conduct ablation studies to investigate the impact of various design choices and hyper-parameters in \PtoP, including w/ or w/o finetuning the local LLM (PEFT), w/ or w/o SMOTE, w/ or w/o RAG, different local/online model backbones, and the number of retrieved entries, $k$, in RAG. 
For the local LLM, we fine-tune two alternative models, MindChat-7B~(MC)~\cite{MindChat7B}, and Chat-GLM-6B (GLM)~\cite{GLM6B, Zeng2022glm}. 
For the online LLM, we use GPT-4 as an ablation while keeping the local LLM as the default DeepSeek. 
Table~\ref{tab:ablation} reports the prediction accuracy of these \PtoP variants in the four \MBTI dimensions. 
Overall, the full version of \PtoP with $k = 5$ achieves the best performance, highlighting the importance of all design options and their synergy. 
When isolating RAG (Variant (e)) or PEFT (Variant (c)/(d)), we observe that RAG contributes more significantly to the overall effectiveness of \PtoP.
As local models, both MC (Variant (g)) and GLM (Variant (h)) underperform DS, where MC is a psychology LLM that slightly outperforms GLM. 
For online LLMs, DS-V3 (Variant (a)) surpasses GPT-4 (Variant (i)) regarding MBTI prediction. We conjecture the reason would be the consistency of LLM type in the dual framework.

\subsubsection{Case Studies}
\label{sec:exp:casestudy}
To further interpret the model’s behavior, we visualize the attention map from the final layer of the local LLM for words with top-ranked weights. 

\begin{figure}[t]
 \centering
	\begin{tabular}[h]{c}
		\vspace{-2ex}
        \subfigure[Prediction: ENTP, Ground Truth: ENTP] {
				\includegraphics[width=1.0\columnwidth]{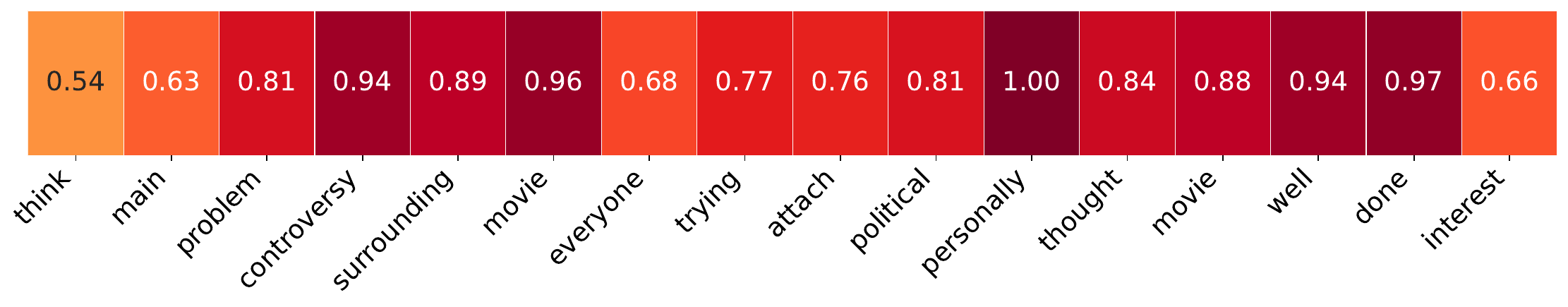}
			\label{fig:T_heatmap}
		} 
        \\
        \subfigure[Prediction: INFP, Ground Truth: INTJ] {
			\includegraphics[width=1.0\columnwidth]{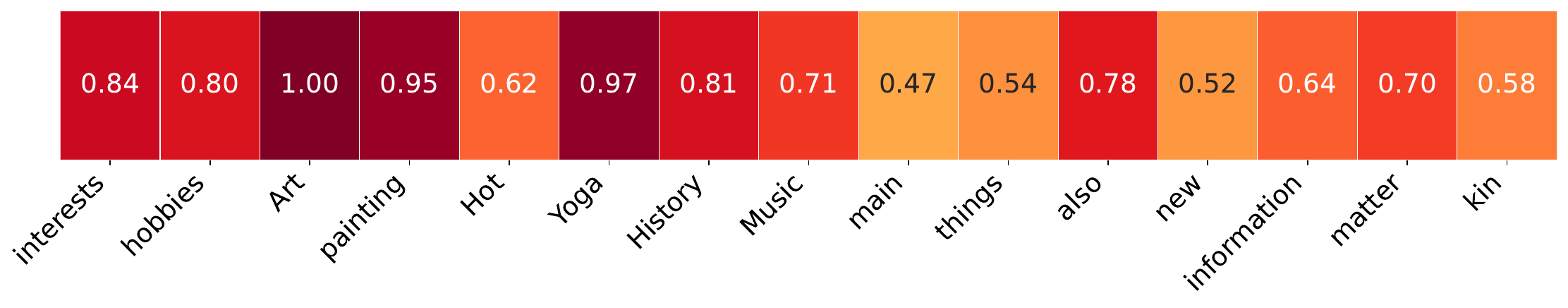}
			\label{fig:F_heatmap}
		}
	\end{tabular}
    \vspace{-4ex}
	\caption{The Attention Maps from Local LLM}
        \label{fig:casestudy}
    \vspace{-1ex}
\end{figure}

Fig.~\ref{fig:T_heatmap} illustrates a well-performed case where both the prediction and the ground-truth label are ENTP. In this case, words such as `problem' (0.81) and `controversy' (0.94) receive high attention weights, closely aligning with the ENTP `Debater' personality traits. Similarly, cognitive-related words like `thought' (0.84) and `think' (0.54) are also assigned prominent weights, reflecting the tendency of ENTP to analyze issues from multiple perspectives and exhibiting the typical characteristic of `thinking for the sake of thinking'—enjoying the process of intellectual exploration. This pattern of attention distribution is consistent with the \MBTI theory, in which Extraverted Intuition (Ne) and Introverted Thinking (Ti) are dominant and auxiliary functions, respectively. This result indicates the model's nuanced understanding of personality features.

Fig.~\ref{fig:T_heatmap} illustrates a misclassified case where the ground-truth is INTJ while \PtoP predicts INFP. We observe that the local LLM excessively focuses on art-related words such as `Art' (1.00) and `painting'  (0.95), reflecting its stereotypical understanding of the outward traits of the INFP `Healer' personality. This misjudgment likely stems from the model’s failure to deeply distinguish the essential cognitive differences between INTJ and INFP types. Key discriminative words such as `strategy' (0.32) and `system' (0.28) receive lower attention, indicating a blind spot in capturing the typical systematic thinking characteristics of INTJ.

\section{Conclusion}
\label{sec:conclusion}
In this paper, we propose PostToPersonality (\PtoP), a novel LLM-based framework predicting \MBTI personality types from social media posts. 
By integrating Retrieval-Augmented Generation (RAG) and LLM fine-tuning with synthetic oversampling, 
\PtoP effectively mitigates the hallucination problem of LLM and addresses the data imbalance issue inherent in \MBTI prediction from real-world data. Experimental results on a real-world dataset demonstrate that \PtoP significantly outperforms 10 existing machine learning and deep learning baselines. 

\section*{Usage of Generative AI}
The source code and datasets of this paper are free of generative AI. Partial texts (excluding equations) in the paper are polished by GPT 4.1 to improve clarity, coherence, and fluency. The submission complies with the ACM policy on the usage of generative AI. 

\bibliographystyle{ACM-Reference-Format}
\balance
\bibliography{ref}

\newpage
\appendix

\section{Implementation details of baselines}
\label{sec:appendix:baselines}
In traditional ML models, the Multi-nomial Naive Bayes computes feature weights by counting term frequencies within posts and combining these with the inverse document frequency, forming a sparse high-dimensional word feature space.
The model utilizes TF-IDF weighted term frequency vectors as input features and employs the default smoothing parameter $\alpha = 1.0$ to balance feature weights.
Logistic Regression (LR) leverages bag-of-words or $n$-gram features as input. 
To mitigate overfitting, the model’s penalty coefficient ($C = 1.0$) and $L2$ regularization type are optimized via grid search.
Support Vector Classifier (SVM) uses normalized TF-IDF vectors as input features. The model uses the RBF kernel, with hyperparameters $C = 10$ and $\gamma = 0.1$ by default. 
XGBoost is configured with a learning rate $\eta = 0.1$, maximum tree depth of 6, and a subsample rate of 0.8. Input features can comprise term frequency statistics as well as embedding features derived from deep learning models. 

The neural network method MLP adopts a hidden layer structure with ReLU activation, matches the input layer dimension to the number of features, uses softmax activation in the output layer, and is trained with an Adam optimizer at a learning rate of 0.001, capable of directly processing word embedding features or splicing them with traditional features. Here, word features are mapped into dense vectors through the embedding layer, preserving lexical semantic information while alleviating the high-dimensional sparsity of traditional one-hot representations.  

Word2Vec uses pre-trained 300-dimensional word vectors and generates sentence-level representations through methods like average pooling or max pooling. 
Among pre-trained language models: e5-base generates 768-dimensional sentence representations using the [CLS] token when fine-tuning on MBTI data, and its word features fuse contextual information through multi-layer Transformer encoders; ModernBERT-base uses a learning rate of 3e-5 and a batch size of 32 during fine-tuning, combined with early stopping to prevent overfitting, and its word feature generation process includes position encoding and multi-head attention mechanisms; DeBERTa-v3-base employs a learning rate of 5e-5 and a layer-wise unfreezing strategy for fine-tuning, enhancing the representational capability of word features through decoupled content and position attention mechanisms.  

These Transformer-based models dynamically generate context-aware word embeddings through self-attention mechanisms to capture long-distance dependencies. In the word feature construction process, the representation of each word is dynamically adjusted based on its contextual sentence, forming fine-grained representations of "polysemy". Specifically, e5-base focuses on efficient embedding generation, optimizing the word feature space through contrastive learning; ModernBERT-base improves the BERT architecture, enhancing the positional sensitivity of word features through rotary position encoding; while DeBERTa-v3-base boosts model performance via decoupled attention mechanisms, enabling word features to model content and relative position information separately. Ultimately, the embedding representations generated by these models can be directly used as input features for downstream classifiers, which contain rich lexical and sentence-level semantic information, significantly improving text classification performance.

\section{Details of the dataset}
\label{sec:appendix:dataset}
\begin{figure}[t]
 \centering
	\begin{tabular}[h]{c}
        \subfigure[Distribution of MBTI Types] {
				\includegraphics[width=0.45\columnwidth]{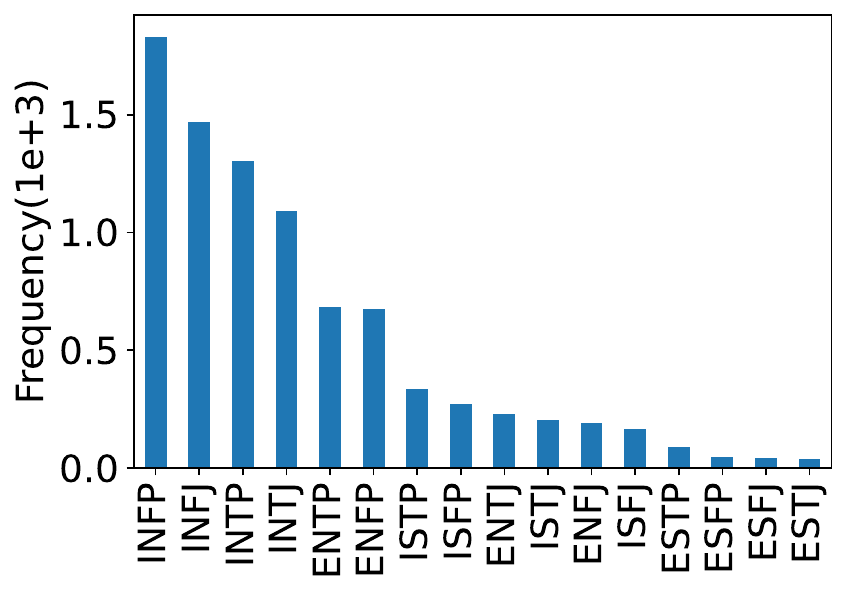}
			\label{fig:personality_types_frequency}
		} 
        \hspace{-1ex}
        \subfigure[Distribution of Post Length] {
			\raisebox{0.32cm}{
                \includegraphics[width=0.45\columnwidth]{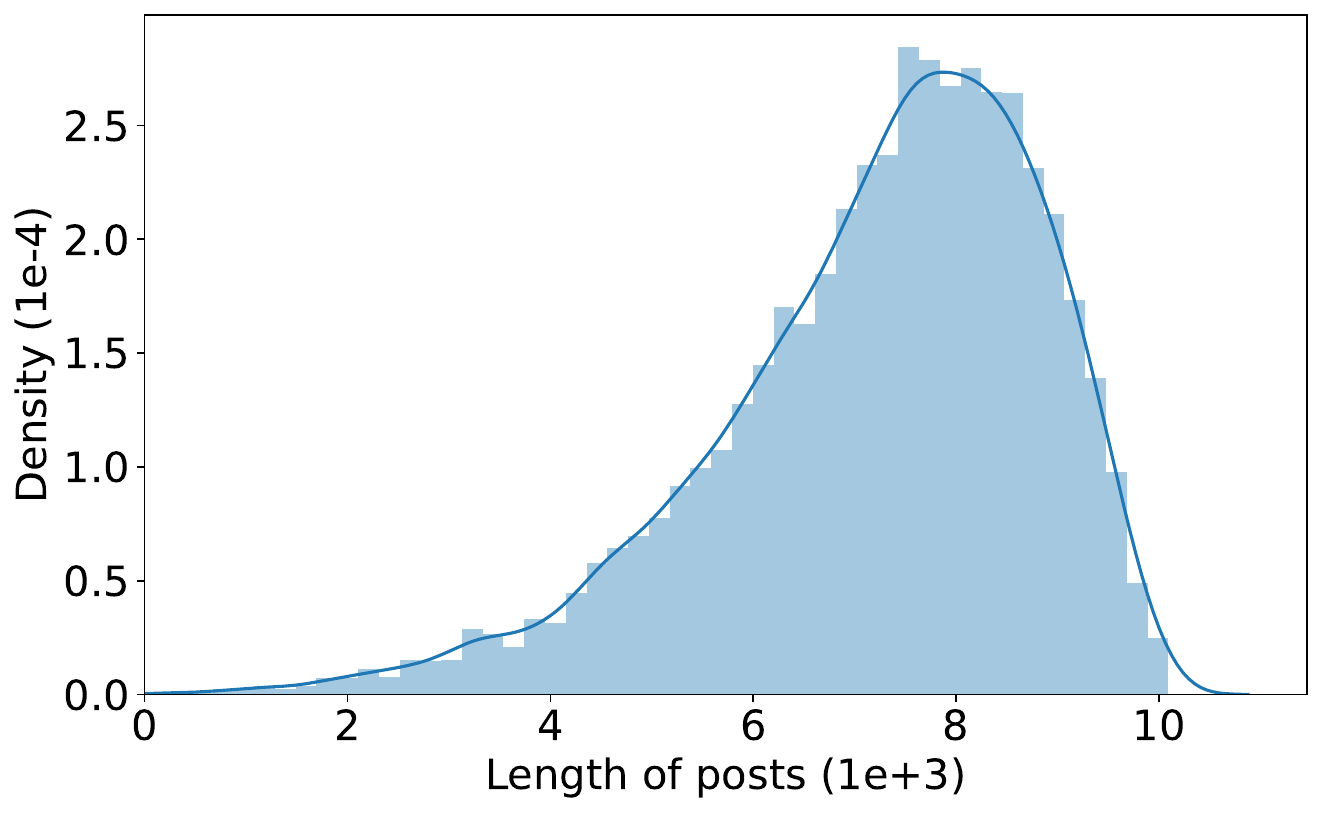}
            }
			\label{fig:post_length_distribution}
		}
	\end{tabular}
    \vspace{-3ex}
	\caption{Statistics of the Dataset}
        \label{fig:distribution}
\end{figure}

Fig.~\ref{fig:distribution} illustrates the statistics of the dataset from \textit{PersonalityCafe}. Fig.~\ref{fig:personality_types_frequency} shows the distribution of the \MBTI types in the dataset, which is highly imbalanced. 
Fig.~\ref{fig:post_length_distribution} shows the distribution of the post length in the dataset with our preprocessing.

\section{The prompt design of P2P}
\label{sec:appendix:prompt}

We introduce the prompt structure employed by  \PtoP. 
%
For feature extraction in the local LLM, the placeholder \colorbox{cyan!60}{<CONTENT>} denotes the social media posts to be analyzed. The instruction explicitly directs the LLM to extract features from the four dimensions of \MBTI.
For online LLM inference, the prompt incorporates three components: \colorbox{cyan!60}{<CONTENT>} identifies the social media posts; \colorbox{yellow}{<FEATURE>} identifies the textual features generated by the local LLM; and \colorbox{green}{<SIM-TEXTS>} identifies the top-$k$ most similar entries retrieved from the vector database, along with their corresponding \MBTI labels. These $k$ entries are arranged in descending order of semantic similarity to serve as reference demonstrations.  
The instruction enforces strict output constraints (`return 4 uppercase letters only') to ensure standardized \MBTI predictions.

\section{Details of case studies}
\label{sec:appendix:casestudies}

We provide the posts and the features generated by the local LLM in the case studies. Posts of the users are concatenated by `|||' into a long text. 

\stitle{Case 1: (Fig.~\ref{fig:T_heatmap})}
(Prediction: ENTP, Ground Truth: ENTP):

\colorbox{cyan!60}{<CONTENT>} 
I personally thought the movie was well done and was interesting to watch. I think the main problem with the controversy surrounding the movie is that everyone is trying to attach their own political...|||Silly Europeans. America is supreme and above criticism :cool:|||Knowing GRRM, The battle at Castle Black is gonna be both epic and heartbreaking :(|||Get ready for the waterworks lol|||:/ That's all I can say about the Mountain vs Viper fight.|||I'm serious considering whether I'm actually an ESTP (i'm probably just gonna go with EXTP for now). And so far, I've been well received in the few contacts I've had with military lifestyle.|||I'm actually the degree for a similar reason as you. I plan on joining the Army after college (18X program if wanna know more) and C.S. degree is more of a fall-back than anything else.|||How come you regret your decision in picking computer science (incidentally, that's the degree that I plan on pursuing in college)?|||One thing I've noticed about all my friends who I think are Se-doms or auxs is they all have steady gazes when their making eye contact. (It almost like their analyzing me :unsure:)  I'm just...|||I've been called weird and awkward but never creepy.|||I'm guessing it would happen randomly for me (that's how I met all my current friends)|||I like it. Interesting and fairly easy to understand|||Ragnar is definitely an EXTP. His desire to explore and find new lands to raid and conquer fit hand-in-hand with a Pe-dom. I'm less sure about whether he is an S or N (I'm leaning towards ENTP due to...|||Just stop cleaning up after their shit. When their parts of house start to mold, they'll get the memo.|||Umm, pirates? (Wait, no that's stupid)  I got nothing.  Anyways, in all seriousness, ESTPs are pretty fun peeps. They're the only ones I know that can keep up (and even surpass) my level of...|||Do you often only use violence for self-defense?   Btw I almost forgot about this thread :P|||So men got banned?|||I can't help that but feel that I would've loved being an explorer of the Americas (a columbus-type person). I think it'd be safe to say that my personal dream job would be that or an...|||Sounds interesting enough. I feel like if we were to fight a big war, our lack of unity would probably cause us to lose faster than anything our enemies could throw at us.|||I'd liked to play as a spellsword or battlemage; roast your enemies from afar with some fire magic and then close in and gut 'em with your handy-dandy sword. :D|||Why not just threaten to beat him up? You are the big brother after all..|||I was about to say the same exact thing. Nothing sets me off more than someone trying to make life decisions for me (I don't mind advice but they need to understand that all it'll be is advice)|||My bad. I didn't really state what I was trying to say very well; my point is that since he is a sociopath at heart, we really can't use MBTI on him since his sociopathy seriously affects how he sees...|||He doesn't have a type. You guys forget that he's a sociopath at his core. He can't be accurately typed b/c he is mentally ill.|||When it comes to politics, I'm pretty much a pragmatist by this point. I started out as a rebel conservative (b/c everyone near me was a liberal) and then started to shift to the right-libertarian...|||Establish fallout shelters across the globe and promptly nuke all major powers. In the post-apocalyptic world, the shelters will act as bases for you to exert influence over the whole world.|||Hi.|||Prevent the enemy from fighting us and lead them in a wild-goose chase. When they're sufficiently exhausted, we unload the wrath of God on the poor bastards and completely eradicate them. Simple....|||What if we just don't show up to the battle?|||Methinks you over-thinking things just a little too much. I agree w/ you that a good of deal of Walt's stunts are unrealistic. But I think it adds to the show, not detracts from it. It helps build...|||What happens when people stop going to college b/c the tuition has become too expensive to pay for? Or the gov't can no longer afford to give student aid for the same reason?   The bubble is going...|||Yeah, I think the wounded are much higher proportionally than previous wars (1 in 8 that are wounded die compared to Vietnam which  was 1 in 3 and WW2 which was 1 in 2)> I guess you can thank better...|||In 2012, more than 25,00 people died in car accidents. By comparison, only 301 military personnel died in Afghanistan in the same year.  Sources: United States Forces casualties in the war in...|||I've heard that you're more likely to die in a car crash than in Afghanistan. (I'll have to pull up the official statistics though)  On-topic, most people should look at community college or...|||Never underestimate it's power...|||I support so much that I'm reviving this 3-month old thread|||Bump   Don't wan't this to be forgotten about|||At least tell me you like SOME type of chocolate?:sad:|||:angry:  How dare you question the deliciousness of hot chocolate, PEASANT!!!!!   Off with your head!!!!!!!!!!!|||There have been calls from many different people (myself included) to create a forum for members of Generation Z (1995-2010/12). I'm posting this as a suggestion to have such a forum created.  ...|||Any proof for this?|||Tie in the main storyline into the guild quests. I like the idea of my guild quest actions altering how the storyline plays out...|||Good luck (and more importantly DON'T volunteer for anything)  BTW what's your MOS?|||Kill her?|||Who are these 'masses' you all keep talking about? I can't seem to find them anywhere...:confused:|||You know, as horrible as sweatshops are, no one ever seems to ask how the people working there feel about it. I'd be willing to guess that working there is a choice of desperation (and like you said...|||Join the Army. That'll make you more interesting (or dead but who cares:tongue:)  All joking aside, I'd say just be willing to do things for the hell of it and over-analyze everything. That's what...|||So, what was the point of your rant again? Or were you just looking for attention?:rolleyes:|||The only time a 'know-it-all' bothers me is when they feel the need to answer a simple question with a half-hour long speech...|||It's just people trying to fit into the stereotype of the NT (you know, borderline sociopathic, a know-it-all, over-theorizing EVERYTHING, that kind of jazz).

\colorbox{yellow}{<FEATURE>} 
 Your personality profile suggests you are someone who enjoys exploring new things and thrives on adventure. You have a strong desire for discovery, a love of taking risks, and a tendency to showcase your abilities among friends. You often appear less concerned with social norms and etiquette, preferring to express your opinions openly. You have a theoretical mindset and frequently demonstrate keen insight into others' motivations and behavioral patterns.

\stitle{Case 2: (Fig.~\ref{fig:F_heatmap})}
(Prediction: INFP, Ground Truth: INTJ):

\colorbox{cyan!60}{<CONTENT>} 
What are your interests and hobbies? Art (painting), Hot Yoga, History, Music <--main things, but also new information (no matter what it is.)  What kind of food do you eat? Anything.   Are you...|||These two are so perfect.|||Je veux le faire, mais je viens juste de commencer à apprendre le français. j'ai utilisé Google Translate pour cela.|||Darn it! Missed it again.:frustrating: What's YW?|||I would like to change my name to Leonine.   Thanks so much!:happy:|||I was in acting for awhile. My favorite part about it (this was before I knew I was an INTJ), was the study of people and relationships. Which in my semi-humble opinion(:proud:) is a big part of it....|||I am the same way, but it's more like awkward and mortified at the same time.|||New understandings Music Sushi Wine Making someone laugh Swimming Painting Learning|||Hello by T.I feat Cee-Lo|||I'm not sure of an addictive personality, but I do obsess over things sometimes.|||I smoke.|||Interesting, never thought about that. :happy:|||I think that article has a little bias or a lack of complete understanding or definition. I don't really have a philosophy of the soul.|||Nonconforming attitude  - I guess. Like @Sanskrit, not sure I know the rules of conformity.  Idealistic  - Generally, yes. Intense curiosity  - Always.     Happy obsession with a hobby or...|||Were you involved in sports or the arts? Kind of, I played everything for like a season :-) Not involved in the arts really.   Dating? Not constantly, but yeah.    Were others intimidated by you?...|||I agree with this. Not quite sure what you meant by the caring coming from the body. For me, mind-caring takes much more effort, especially when it is something that I think is of a trivial nature....|||Love some of it.|||Would you ever adopt a child? Possibly. Would you ever cheat on your partner? I'd do my best not to.   Would you ever take a bullet for someone else? Yes.     Would you ever slap...|||Just English, trying to learn French.|||Right now I paint. I want to study more about art and I want to learn more languages (trying to learn french now).|||Not really a style per se, but my staples are: Summer/Spring: Dresses, shorts, tanks Fall/Winter: Big Chunky sweaters. cardigans, jeans, boots, white/grey tees and tanks||| We are like children We;re painted on canvases Picking up shades as we go We start off with gesso brushed on by people we know Watch your step as...|||Wine, Champagne, Cigarettes, Painting , Water (bath, swimming,etc.)|||If I remember this correctly: I had an easy bake oven that I used to bake things in. I played with dolls (always did their hair). I use to battle it out with action figures, draw A LOT. And I had a...|||My opinion-It is also irrational to claim to know the answer. Both sides of the coin are a belief, based on the information and understanding you have within your grasp.   Although I do think lib...|||Thanks! :happy:    Not sure if this was supposed to be positive or negative, but thank you for your feedback.     That is what popped in my mind when I completed it, and I thought it fit....|||Thanks! :blushed: It's 20 x 20 inches. The white in the foreground is simply the canvas while the circle in the background is white paint.   It's a figurative painting. I'm glad you like it. Hooray...|||That's beautiful. Thanks for sharing. :blushed:|||These  statements aren't correct. He was speaking to the Pharisees and Sadducees of the time who were the upholders of the oral law and written law.   But when he saw many of the Pharisees and...|||Constantly.|||No, not in real life, over the internet. I'm even worse in real life:frustrating:. Yeah, I try to be as concise as possible, it doesn't seem to work. Ah well, you live and you learn.|||It seems to me, that when I try to make a point I'm either not reading disagreements/debates correctly or I'm not getting across what I'm trying to say.   It seems like everything gets responded to...|||Thanks, I think I will go visit them for a bit.:happy:|||Sometimes I think I will never be able to communicate effectively, therefore I will never be able to be understood. :sad:|||Not at all. I'm not sure why this is though. Possibly because what I choose to listen to is very personal for me, so I have no desire to know what the artist's intention is/was (as horrible as that...|||I agree. MirrorSmile I don't know if this helps but if I'm not giving people the reaction they want, I ask them what do you want me to do?(Not in a negative way) This sometimes makes what they're...|||I really don't. I could be wrong though.  Maybe by making an effort to talk to someone, but sometimes I do this even when I'm not flirting.|||That would be unpopular for me, I don't think anyone deserves death or to be made a science experiment. (But I'm also against the death penalty).   Here's mine:  1)Abandon left wing/right wing...|||DesertWind Really liked that.|||Thanks! Thank you for starting this thread. :-)||| Yay! Paper!|||1)Probably get a little defensive (if I was still in my feelings) 2)Calm down 3) Acknowledge, Admit, and Fix|||I don't know about the Fi thing but I was reading a convo on another forum that went like this:  Title: I need help with my INTJ arguing. ENFP(I think): INTJs should not argue or debate with...|||The conversations between T's and F's can be quite interesting.:frustrating::laughing:|||I didn't think it was that bad:unsure:|||What's wrong?|||I think this is untrue. There were quite a few people in the thread who were contributing to the conversation, but it never went anywhere.  I also think, if your goal in the conversation is to...|||I apologize, I misunderstood you.

\colorbox{yellow}{<FEATURE>} 
Emotional intensity; Interests: Art, yoga, history, music, painting, learning new languages; Personality traits: Stubbornness, passion; Habits: Wine, champagne, cigarettes

\end{document}